\documentclass[10pt,twocolumn,letterpaper]{article}

\usepackage{3dv}
\usepackage{times}
\usepackage{epsfig}
\usepackage{graphicx}
\usepackage{amsmath}
\usepackage{amssymb}
\usepackage{esvect}
\usepackage{cleveref}
\usepackage{booktabs}
\usepackage{subcaption}
\usepackage{xcolor}
\usepackage{duckuments}
\usepackage{placeins}
\usepackage[font=small,labelfont=bf,tableposition=top]{caption}

\newcommand*\rfrac[2]{{}^{#1}\!/_{#2}}%running fraction with slash - requires math mode.

\threedvfinalcopy % *** Uncomment this line for the final submission

\def\threedvPaperID{209} % *** Enter the 3DV Paper ID here

\ifthreedvfinal\fi
\begin{document}

%%%%%%%%% TITLE
\title{TermiNeRF: Ray Termination Prediction for Efficient Neural Rendering}

\author{Martin Piala\\
Imperial College London\\
{\tt\small pialamar@gmail.com}
\and
Ronald Clark\\
Imperial College London\\
{\tt\small ronald.clark@imperial.ac.uk}
}

\makeatletter
\let\@oldmaketitle\@maketitle% Store \@maketitle
\renewcommand{\@maketitle}{\@oldmaketitle% Update \@maketitle to insert...
% \begin{figure*}[ht]
  \def \factor {0.136}
  \def \vertspace {0cm}
  \def \vertspacebig {0.2cm}
  \def \horizontalspace {0.1cm}

%   \centering
  \begin{tabular}{c@{\hspace{\horizontalspace}}c@{\hspace{\horizontalspace}}c@{\hspace{\horizontalspace}}c@{\hspace{\horizontalspace}}c@{\hspace{\horizontalspace}}c@{\hspace{\horizontalspace}}c}
  \includegraphics[width=\factor\linewidth]{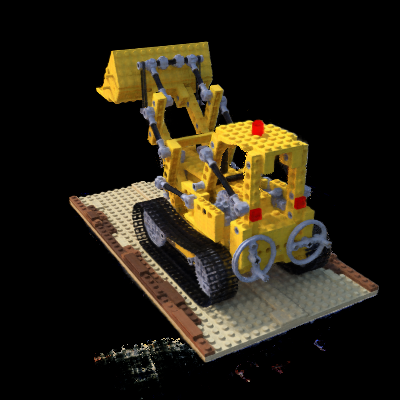} &
  \includegraphics[width=\factor\linewidth]{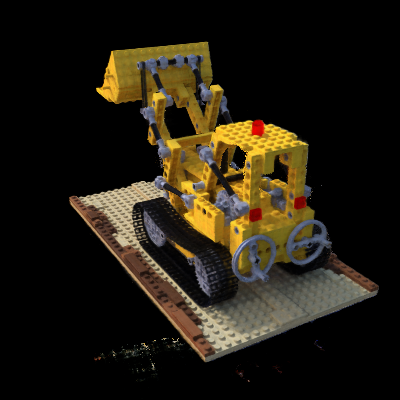} &
  \includegraphics[width=\factor\linewidth]{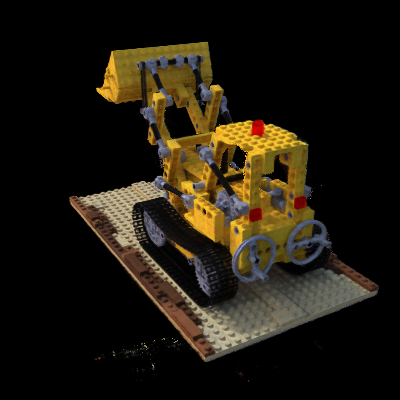} &
  \includegraphics[width=\factor\linewidth]{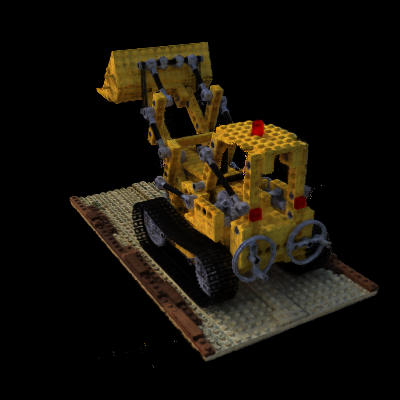} &
  \includegraphics[width=\factor\linewidth]{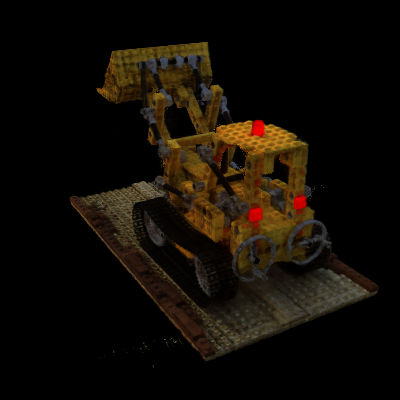} &
  \includegraphics[width=\factor\linewidth]{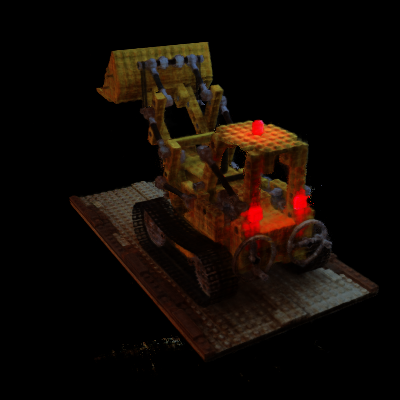} &
  \includegraphics[width=\factor\linewidth]{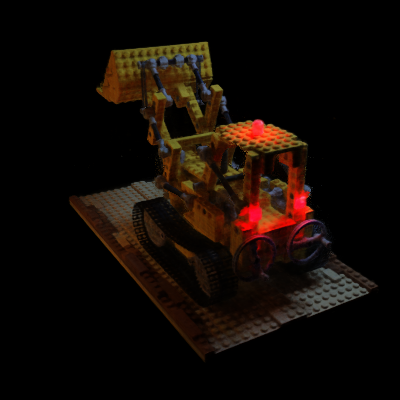}\\
  \end{tabular}
  \captionof{figure}{Accurate sampling allows for fast fine-tuning to a modified scene. This scene was fine-tuned up to the penultimate step in only three minutes on GTX1080.
  Any residual noise is removed from the last render after 1h of additional training.}
  % \caption{Banner}
  \label{fig: banner}
% \end{figure*}
}
\makeatother

\maketitle

\thispagestyle{empty}

% However, as neural fields are implicit models they need to be queried at multiple points along each ray in order to render an image, resulting in very slow rendering times.
%%%%%%%%% ABSTRACT
\begin{abstract}
Volume rendering using neural fields has shown great promise in capturing and synthesizing novel views of 3D scenes.
However, this type of approach requires querying the volume network at multiple points along each viewing ray in order to render an image, resulting in very slow rendering times.
In this paper, we present a method that overcomes this limitation by learning a direct mapping from camera rays to locations along the ray that are most likely to influence the pixel's final appearance. Using this approach we are able to render, train and fine-tune a volumetrically-rendered neural field model an order of magnitude faster than standard approaches. Unlike existing methods, our approach works with general volumes and can be trained end-to-end.
\end{abstract}

%%%%%%%%% BODY TEXT

\section{Introduction}

The idea of using neural networks to represent a 3D scene as a continuous neural field has recently emerged as a promising way to represent and render 3D scenes \cite{Park_2019_CVPR, OccupancyNetworks, mildenhall2020nerf}. In general, this type of representation uses a neural network to learn a function that maps coordinates to optical properties of the scene, such as color and opacity, at each point in space. Unlike any of the traditional representations such as voxel grids, meshes or surfels \cite{pfister2000surfels}, neural fields do not require any discretization of the scene.

While this type of representation captures scenes at very high quality, a major limitation is that it takes a long time to train the model and render images. This is because the neural network representing the volume has to be queried along all the viewing rays. For an image of width
\(w\), and height \(h\), and \(n\) depth samples, the rendering requires \(\Theta(nhw)\) network forward passes.
With NeRF, it can take up to 30 seconds to render a \(800 \times 800\) image on a high-end GPU.

\begin{figure}[t!]
    \centering
    \includegraphics[width=0.9\columnwidth]{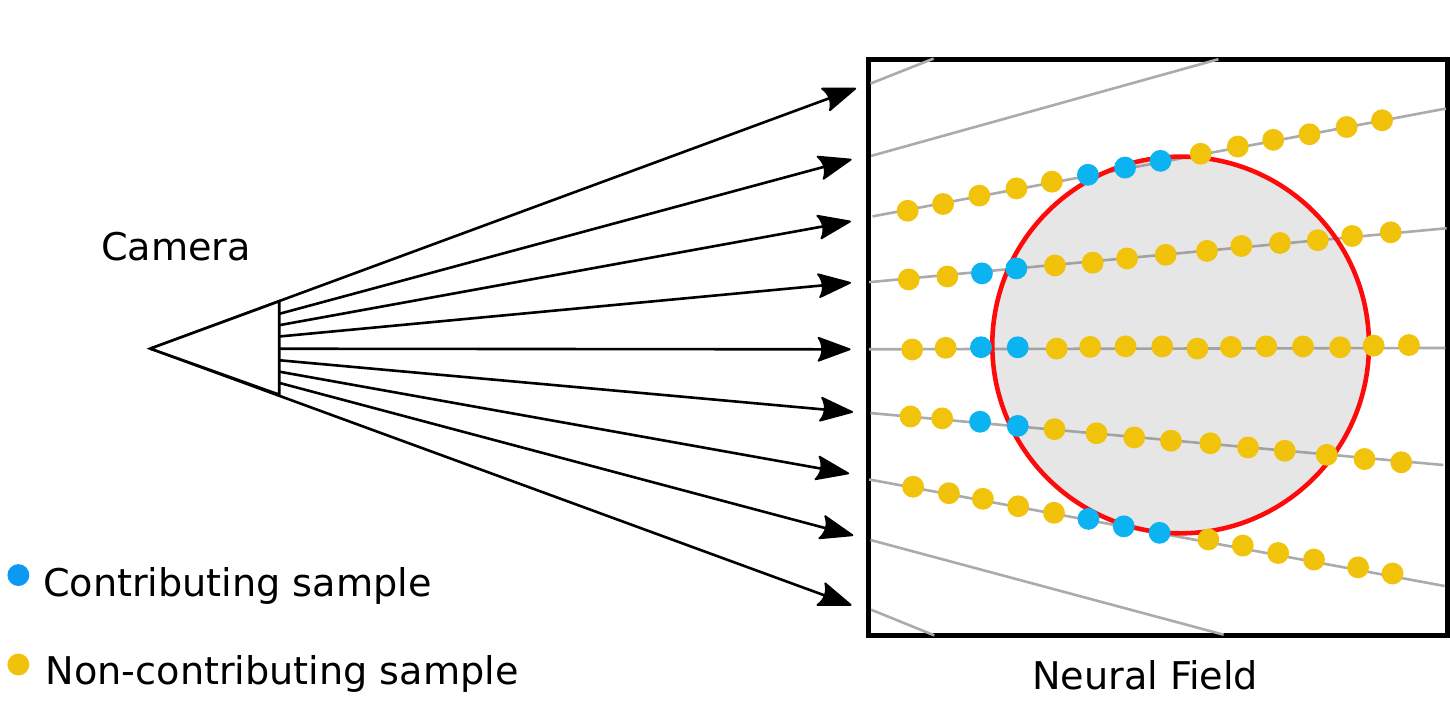}
    \caption{Of the many points sampled along each viewing ray, only a very small fraction contribute to the final color of the pixel. Our approach, called TermiNeRF, efficiently predicts the most important samples, significantly speeding up the rendering.}
    \label{fig:my_label}
\end{figure}

In this paper, we propose a more efficient way to render (and train) a neural field model. Our approach works by jointly training a \textit{sampling network} along with a \textit{color network}, with the collective name \textit{TermiNeRF}. The sampling network estimates where along the viewing direction the surfaces lie, thereby allowing the color network to be sampled much more efficiently, reducing the rendering time to $\leq 1$s. %We take this idea further and propose to learn a mapping directly from ray to radiance, reducing the number of network evaluations to $O(hw)$.
We show how this approach can be applied to quickly edit a scene by rapidly learning changes to lighting or materials for a scene.

\begin{figure*}[h!]
  \centering
  \includegraphics[width=\textwidth]{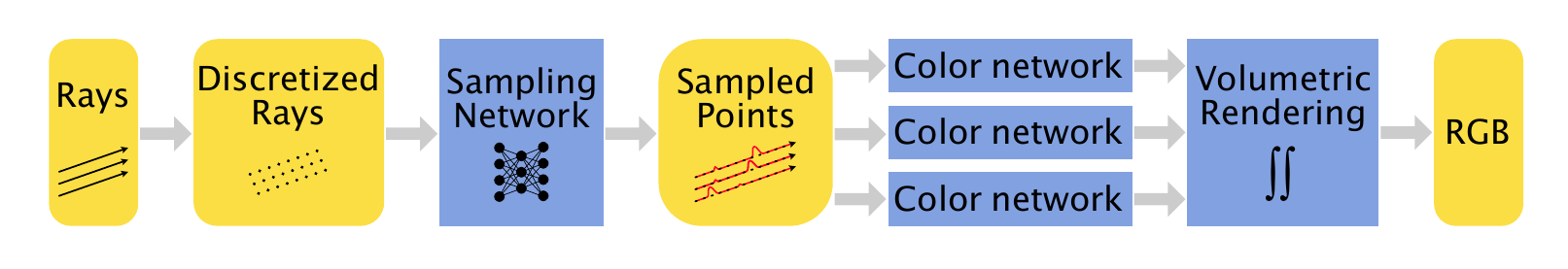}
  \caption{
  Rendering Pipeline: Rays are first discretized and each is evaluated by the sampling network. The sampling network then estimates the distribution of ideal sampling locations along the ray. Ray points drawn stochastically according to this distribution are then separately evaluated by the color network. The final color is aggregated via volumetric rendering.}
  \label{fig: method rendering pipeline}
\end{figure*}

\section{Related Work}

In this section, we give an overview of the related work on raymarching neural network-based 3D scenes and existing attempts to make these approaches more efficient.

\subsection{Raymarching Neural fields}
Using a neural network that represents a continuous scalar or vector function through space has recently emerged as a promising way to parameterize 3D objects and scenes for rendering. Such models go by various names, including ``neural coordinate-based representations" or ``neural implicit models", but perhaps the most appropriate and general term is ``neural fields" \cite{hinton2021represent}.

The first approaches in this direction were focused on representing geometric shapes as distance fields and include methods such as DeepSDF \cite{Park_2019_CVPR}, Occupancy Networks \cite{OccupancyNetworks}, and IM-Net \cite{implicit_decoder}. These approaches are generally trained using a sampled 3D point cloud of the shape as supervision. DVR \cite{DVR} introduced a ``differentiable'' volumetric rendering method to train a neural occupancy field using only 2D images as supervision. The main disadvantages of DVR is that it is only able to reconstruct solid surfaces, and view-dependent effects caused by reflections and specularities cannot be captured.

In contrast, neural radiance fields (NeRF) \cite{mildenhall2020nerf} use a neural network to parametrize the color and opacity of each point in space, coupled with volumetric raymarching, to render novel views. The main disadvantage of NeRF is that the volumetric rendering requires sampling the network at multiple positions along a ray projected through each pixel, leading to a prohibitively high computational cost.

More recent works have tried to deal with images captured in uncontrolled conditions with transient distractors and varying lighting conditions \cite{martinbrualla2020nerfw}. Other works have looked at how the scene can be decomposed into components such as reflectance and incoming radiance so that the scene can be re-lit with new lighting.

PixelNeRF \cite{yu2021pixelnerf} and MVSNeRF \cite{chen2021mvsnerf} proposes a generalized NeRF approach conditioned on image inputs, enabling memory-efficient rendering of various scenes with a small training overhead.

\subsection{Efficient neural radiance fields}

In order to improve the efficiency of the volumetric rendering of neural radiance fields, two main types of approaches exist. The first type tries to reduce the number of samples taken along each ray by ``intelligently'' picking sample locations \cite{neff2021donerf,arandjelovic2021nerf}. \cite{deng2021depth} utilizes depth supervision for more efficient sample placement during the training. DONeRF \cite{neff2021donerf} takes this idea further and learns a separate depth network that is trained using ground-truth depth maps to predict possible surface locations. This allows them to sample more efficiently in the region near the surface.

\cite{arandjelovic2021nerf} uses a transformer or an MLPMixer \cite{tolstikhin2021mlp} to predict sampling locations from an input ray. Although being end-to-end trainable, this method only achieves $25\%$ speedup with a significant degradation in quality.

The second approach type tries to improve the efficiency of sampling the field, thus reducing the time taken for each evaluation of the scene model. DeRF \cite{Rebain_2021_CVPR} decomposes the space into a number of regions and uses a smaller MLP for learning the appearance of each region, which significantly reduces the rendering time. KiloNeRF \cite{kilonerf} takes this idea to the extreme and uses a very large collection of many small MLPs. FastNeRF \cite{garbin2021fastnerf} factorizes NeRF into two MLPs and speeds up the rendering process via caching.

\begin{table}[b]
  \caption{Comparison of efficient neural field rendering approaches. Speedup is relative to the standard NeRF model (1x).}
  \centering
  \begin{tabular}{l|l|l|l}
              & Speedup & End-to-End                          & Requires depth                      \\ \hline \hline
  DONeRF      & 48x   & {\color[HTML]{FE0000} \textbf{No}}  & {\color[HTML]{FE0000} \textbf{Yes}} \\
  DeRF        & 3x    & {\color[HTML]{FE0000} \textbf{No}}  & {\color[HTML]{32CB00} \textbf{No}}  \\
  KiloNeRF    & 600x  & {\color[HTML]{FE0000} \textbf{No}} & {\color[HTML]{32CB00} \textbf{No}}  \\
  AutoInt     & 10x   & {\color[HTML]{32CB00} \textbf{Yes}} & {\color[HTML]{32CB00} \textbf{No}}  \\
  PlenOctrees & 3000x & {\color[HTML]{FE0000} \textbf{No}}  & {\color[HTML]{32CB00} \textbf{No}}  \\
  Ours        & 14x   & {\color[HTML]{32CB00} \textbf{Yes}} & {\color[HTML]{32CB00} \textbf{No}}
  \end{tabular}
  \end{table}

\cite{yu2021plenoctrees} replaces the MLP with a ``Plen'' Octree that stores the direction-dependent radiance at each position. This octree is orders of magnitude faster to sample than an MLP. Instead of learning color and opacity at individual locations and integrating them during volumetric raymarching, Autoint \cite{autoint} learns a network that can predict the integrated color of ray segments. In so doing, it amortizes the integration and makes rendering noticeably faster.

\section{Method}

In this section, we introduce our method for fast rendering and, by extension, training of neural radiance fields models. The key insight of our method is that the only samples that contribute to the final color of a ray are ones which have non-zero opacity (i.e., samples near surfaces or participating media) and are not already occluded. Thus, our method works by learning a distribution along each ray, via a \textit{sampling network}, that represents the probability of the ray encountering a surface or some participating media at each point. With the information about the distribution of the volume along the ray, we were able to make the rendering pipeline (depicted on \cref{fig: method rendering pipeline}) more efficient by skipping points that lie in free space.

\subsection{Network models}
In this section we describe the two network models used in our approach - the color and sampling network. The color network is a replica of NeRF's \cite{mildenhall2020nerf} fine network and models color as a function of position. \label{sampling network}
Splitting up a ray into disjunct segments, or \textit{bins}, the \textit{sampling network} estimates the corresponding weights to sample points for further evaluation by the \textit{color network}.
With a suitable ray representation, the network is able to do such prediction using only one evaluation per ray, speeding up the rendering process. The weights \(w\) predicted by the network are normalized such that \(\hat{w}_i = \rfrac{w_i}{\sum_{j = 0}^N w_j}\). We observed that a sampling network of the same size and architecture as a NeRF network (8 hidden layers with 256 units and a skip connection to the fifth layer) has enough capacity to learn good samples.

\subsection{Ray parameterisation} \label{sec: ray parameterisation}
A single ray can be effectively parametrized by infinitely many combinations of the origin and direction \((O, \vv{d})\). The sampling network would not only have to predict the depth but also solve the ambiguity of the ray parameterization. To lift the computational burden from the network, we need to add a structure to the parametrizations and, with additional constraints, establish a bijection between rays and their representations. With an additional assumption that the initial point \(O\) of each ray always lies outside of the sphere circumscribing the scene, the exact origin of the ray is not relevant anymore, and we can choose to replace it with any point on the ray or not to store it altogether.

\textbf{Sphere intersection:} \label{sec: sphere intersection}
One such way to unify the ray parametrizations is to represent it by two points of intersection with a sphere circumscribing the scene \cite{neff2021donerf} and the direction by the relative order of the points.
One issue of such representation is that the line segments delimited by the intersection points are of a different length, depending on the distance of a ray from the center of the sphere. Uneven segment lengths create additional computational pressure on the network.

% \begin{figure}[h]
%   \centering
%   \includegraphics[width=0.8\linewidth]{3 Method/figures/two_points_on_sphere.png}
%   \caption{Ray representation: Sphere intersection}
%   \label{method sphere intersection representation}
% \end{figure}

\textbf{Constant length segment:} \label{sec: constant length segment}
To rectify the aforementioned issue, we can enforce the user-defined length \(\ell\) of each line segment. This line segment is defined by two points \((A, B)\), such that \(|B - A| = \ell\). The direction of the ray is given by the relative order of the points \(\vv{d} = B -  A\). To ensure the bijection between the points and rays, we further require the midpoint of the segment to be the line’s closest point to the origin of the scene \(proj_{\vv{AB}}O = \frac{A + B}{2}\). Following the length of the ray from \cite{mildenhall2020nerf}, we set \(\ell = 4.0\).

% \begin{figure}[h]
%   \centering
%   \includegraphics[width=0.8\linewidth]{3 Method/figures/two_points_equidistant.png}
%   \caption{Ray representation: Segment of constant length}
%   \label{method constant segment representation}
% \end{figure}

\textbf{Additional samples:}
With the above parametrizations, there is still a considerable amount of complexity left for the network as it would have to solve for intersections in between the two potentially distant points. Similarly to work by Neff et al. \cite{neff2021donerf} we propose additional samples in between the already determined points from the previous sections. With the assumption that the majority of the volume is being located in the middle of the scene, we propose \textit{centred-logarithmic} sampling, where the points are logarithmically sampled from the middle of the determined line segment towards the edges. The possible boundaries of the segment are described in \cref{sec: sphere intersection,sec: constant length segment}. Mathematically speaking, if the two boundary points are \(A\) and \(B\), the placed samples are:
\begin{equation}
  \left\{
  A,
  A + s_2 \cdot \vv{d},
  \cdots,
  A + s_{N-1} \cdot \vv{d},
  B
  \right\}
\end{equation}

where \(s\) is formed by the concatenation of \(l\) and \(u\):
\begin{equation}
  l_i = 1 - 2^{\frac{1-i}{N/2 - 1}} ; \;\; i \in \{1, ..., N/2 - 1\}
  % l_i = 0.5 - \frac{2^{\frac{N/2 - i}{N/2 - 1}} - 1}{2} ; \;\; i \in \{1, ..., N/2 - 1\}
\end{equation}
\begin{equation}
  u_j = 2^\frac{j - N/2}{N/2 - 1}; \;\; j \in \{1, ..., N/2\}
  % u_i = 0.5 + \frac{2^\frac{i-1}{N/2 - 1} - 1}{2}; \;\; i \in \{1, ..., N/2\}
\end{equation}

where \(\vv{d} = B - A\) is the ray direction.

The additional points form a neat definition of the ray bin boundaries we wish to predict the weights for. The last bin spans from the last boundary point up to infinity. Using points as bins boundaries additionally means that the two edge points \(A, B\) has to lie outside of the rendered object.

\textbf{Representation comparison:}
\begin{figure}[h]
  \centering
  \includegraphics[width=0.92\linewidth]{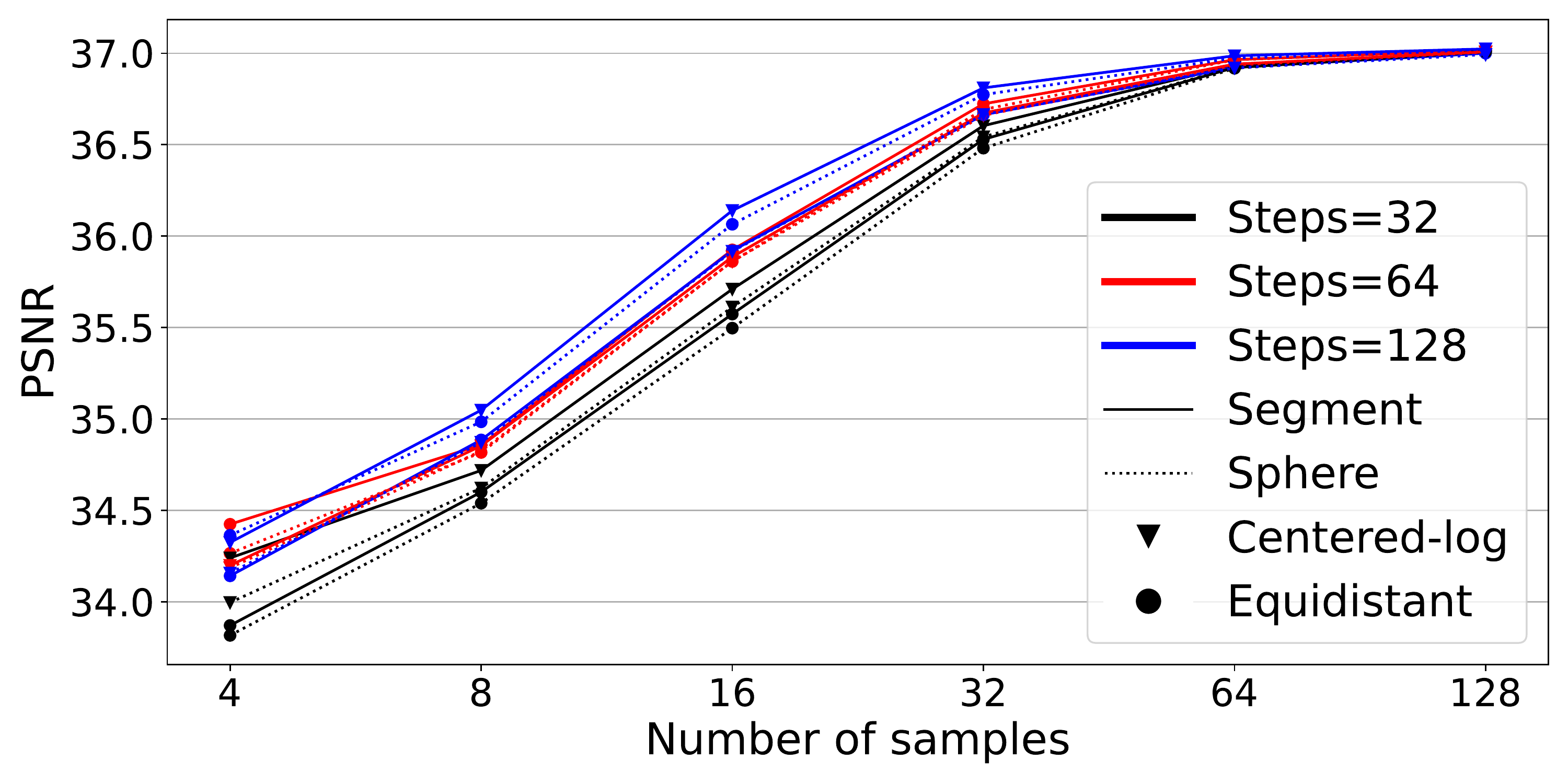}
  \caption{Ray Parametrization Comparison: PSNR as measured on the versatile Lego dataset. For each representation (\cref{sec: ray parameterisation}), we trained a separate sampling network for 85 epochs and selected the checkpoint with the lowest validation loss. We experimented with different numbers of rendering samples, and the dominance of segment + centered-logarithmic samples across the majority of the spectrum is noticeable.}
  \label{fig: method ray parameterisation comparison}
\end{figure}
We have experimented with various ray representations and combinations of boundary points as well as additional samples. For completeness, to centered-logarithmic sampling we also added equidistant samples.
The overview can be found on \cref{fig: method ray parameterisation comparison}. As the higher number of bins in combination with centered-logarithmic sampling and segment ray representation are favored, we will keep this set-up for all other experiments unless specified otherwise.

\subsection{Sampling distribution learning} \label{sec: depth dataset}

NeRF consists of two jointly-trained, coarse and fine, networks. The NeRF model approximates the 5D function \(F(x, \vv{d}) \rightarrow (c, \sigma)\) and estimates the color \(c\) and the optical density \(\sigma\) of each point \(x\) in the corresponding viewing direction \(\vv{d}\). To train the sampling network, ground truth weights are obtained from the second, fine, network in a similar fashion they are obtained from the coarse network in the original rendering pipeline:
\begin{equation}
  w_i = \exp\left( - \sum_{j=1}^{i-1} \sigma_j \delta_j \right) (1 - \exp(-\sigma_i \delta_i))
\end{equation}
where \(\delta_i = z_{i+1} - z_i\) is the distance between two neighbouring samples.
For each ray, NeRF is evaluated 64+128 times, therefore 196 tuples \((z_i, w_i)\) are recorded. The \(z_i = \frac{|O - x_i|}{|\vv{d}|}\) are the z-values capturing the distance of each point from the origin of the ray.

The sampling network estimates \(N\) weights corresponding to the \(N\) bins along the ray.
As label weights are obtained from NeRF, inherently, there is a certain amount of noise involved. Therefore we perform \textbf{Gaussian blurring} of the ground truth weights along the ray with a fixed window size, where the window size covers all samples within the given Euclidean distance from the center of the window.

\begin{figure}[h]
  \centering
  \includegraphics[width=0.94\linewidth]{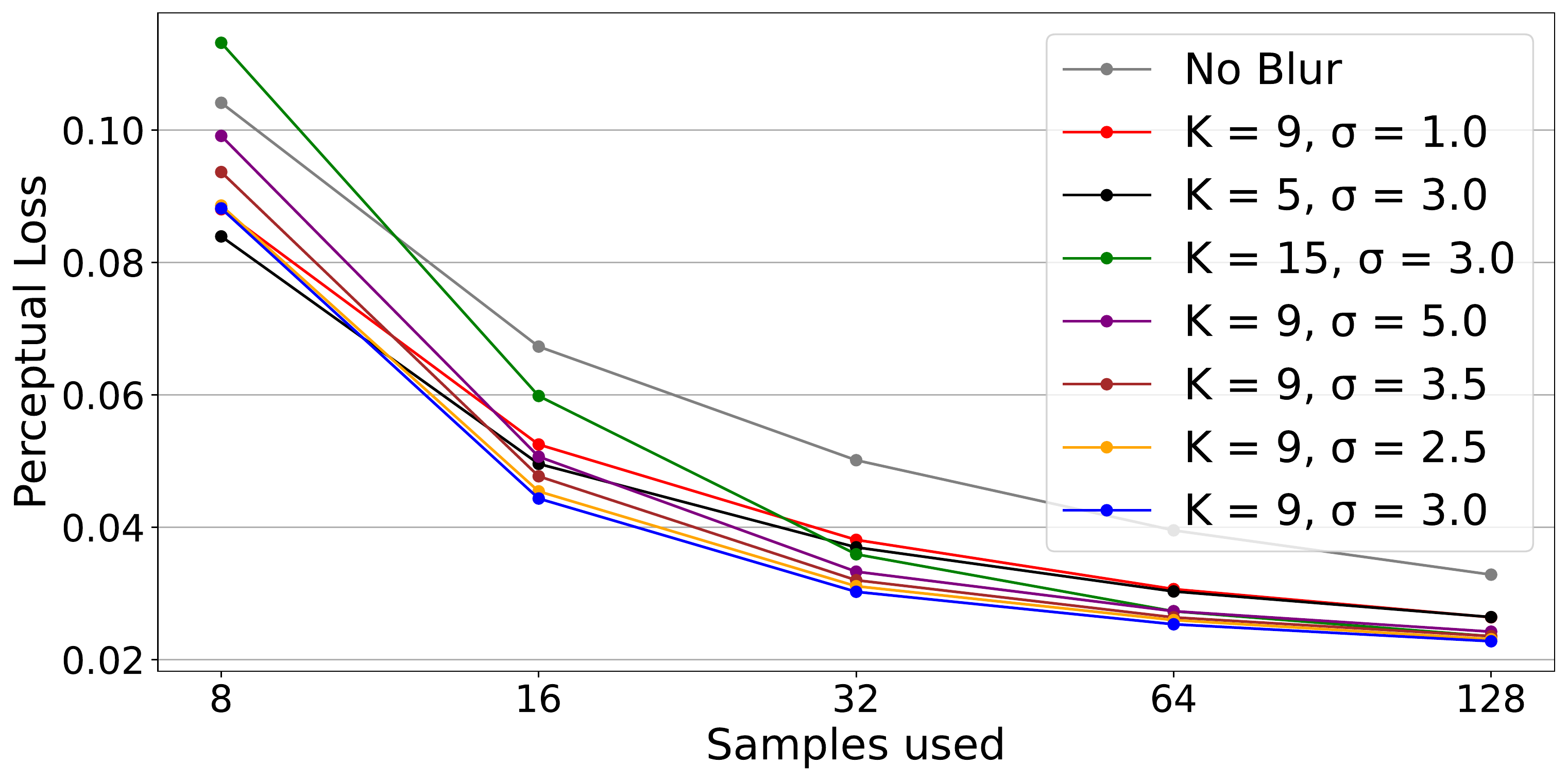}
  \caption{
    Ablation study: Gaussian blur on the Lego dataset. Blur significantly helps to lower the perceptual loss LPIPS (AlexNet). The ideal values we found are \(K = 9, \sigma = 3.0\).
    The Euclidean distance the window covers is calculated for the Lego dataset as \(d = \frac{K}{N_{\textit{samples}}} * \ell = \frac{K}{128} * 4.0\). Note that the number of elements inside the window of a constant size depends on the local density of samples. The uncertain number of elements inside each window could pose an issue if such computations are performed on modern GPUs. Therefore an additional distribution matching step can be performed before the blur to place samples evenly along the ray.}
  \label{fig: ablation gaussian blur}
\end{figure}

Furthermore, it is not guaranteed that the ground truth weights will be defined in the desired z-values (bin boundaries). Therefore, in order to find the correct label weights for our bins, a \textbf{distribution matching} is necessary as the training would not be possible without correctly placed label weights along the ray.

Given a distribution defined as \(N\) z-value and weight pairs \((z_i, w_i), \forall \; 1 \leq i \leq N\), we wish to find the new label weights \(\hat{w}_j\) corresponding with the desired bins.
% defined by boundary points \(\hat{z}_j, \forall \; 1 \leq j \leq M\).
For the correct behavior of the network, certain properties of the distributions have to be preserved:
\begin{itemize}
  \item Any peak in the original distribution, however narrow, has to appear in the resampled variant, too. This problem becomes especially pronounced when the old and new z-values are not spaced out evenly.
  \item Without a further subdivision, two segments of the same size on the ray are equally important for rendering if their largest weights are equal, irrespective of the number of samples in each segment.
\end{itemize}

\begin{figure}[b]
  \centering
  \includegraphics[width=\linewidth]{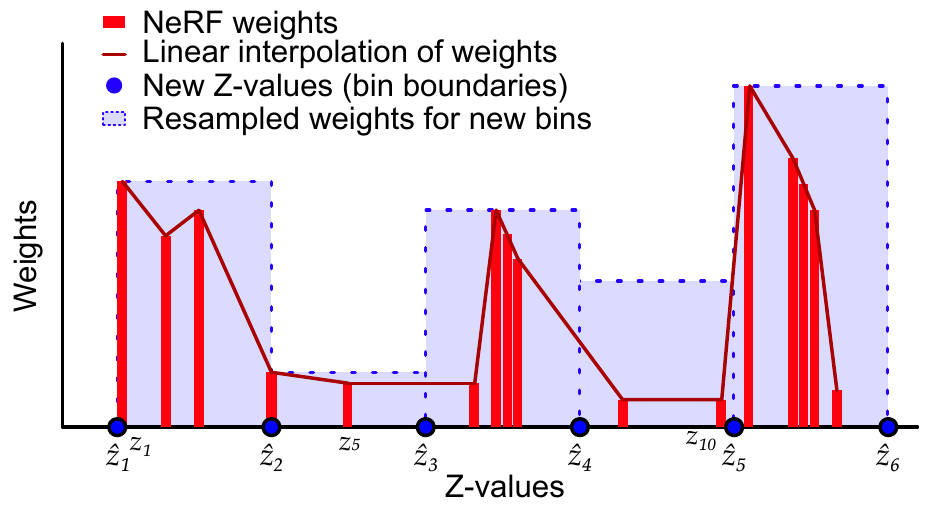}
  \caption{Distribution matching visualization. The weight of each bin is the maximum of all the original weights located inside the bin. For illustration purposes the NeRF weights are depicted as a thin bar, even though in the original NeRF model the magnitude determines the sampling weight of the whole region until the next z-value.}
  \label{fig: method distribution matching}
\end{figure}

To maintain these properties we propose \textit{max-resampling}.
Assuming a bin \(b_j\) is delimited by two consecutive z-values \([\hat{z}_j, \hat{z}_{j+1}]\), the weight corresponding to the \(b_j\) is chosen as the largest weight within the bin boundaries from the original distribution as depicted on \cref{fig: method distribution matching}. Mathematically, for each \(b_j\), the weight
\begin{equation}
  \hat{w}_j = \max \left( \{w_i |\; \hat{z}_j \leq z_i \leq \hat{z}_{j+1} \} \cup \{w'_j, w'_{j+1}\} \right)
\end{equation}
  is chosen, where \(w'_j\) is linear interpolation of the weights from the old distribution for the z-value \(\hat{z}_j\).
As the sampling network output is normalized, blurred and distribution-matched weights also have to be normalized. Then, using an MSE loss function, the resampled and normalized weights distribution can be used as the supervision for the sampling network.

% The color network is initialized with the parameters of the pre-trained NeRF fine network, and further fine-tuned on samples provided by the sampling network.

% \begin{figure}[h]
%   \centering
%   \includegraphics[width=0.95\linewidth]{example-image-duck}
%   \caption{Blur values comparison}
%   \label{method distribution matching}
% \end{figure}

% \subsubsection{Objective function}

% The loss function consists of a simple MSE on weights. \(w\) being the predicted weights and \(\hat{w}\) being the resampled blurred ground truth weights:
% \begin{equation}
%   \ell_{\hat{w}} (w) = \MSE(w, \hat{w})
% \end{equation}

\section{Experiments}
Our primary motivation was to extend NeRF \cite{mildenhall2020nerf} and make the sampling more efficient. We therefore compare our approach with the original implementation of NeRF on multiple datasets that contain both coarse and fine structures, as well as semi-transparent volumes demonstrating the power of multilayer depth prediction. For each dataset we record the average PSNR and LPIPS \cite{zhang2018perceptual} based on AlexNet \cite{krizhevsky2012imagenet} over 200 testing images.

Furthermore, we demonstrate the potential of our set-up to train the sampling network jointly with the color network, leading to more accurate depth and color predictions.

Lastly, we demonstrate how the sampling network can be utilized in more efficient retraining and fine-tuning of the color prediction network for scene modifications.

\subsection{Datasets}
For the most part we use the original NeRF \cite{mildenhall2020nerf} datasets \textbf{Lego}, \textbf{Mic}, \textbf{Ship} and \textbf{Drums} for testing and comparative analysis. We also add a few additional scenes to test the performance of our model in specific scenarios.
The first additional scene, called \textbf{Radiometer}, contains multiple layers of volume, as it shows an object inside a glass enclosure.

Additionally, we generate a modified version of the Lego and Ship datasets: \textbf{Lego Night} and \textbf{Ship Night}, which contain the same geometry as the originals but are rendered under darker night-like conditions. These are to demonstrate the fine-tuning ability of our method.

\subsection{Faster Rerendering novel views}
The purpose of this experiment is to show that with the use of the sampling network, we can accurately decrease the necessary number of color network evaluations and render the scenes faster.

\textbf{Experiment set up:}
For each dataset, we compare the best achieved NeRF results with our method.
We report values for multiple rendering time brackets, as there is a natural trade-off between the rendering speed and the image quality as we tweak the number of rendering samples.

After 100 epochs of sampling network training, we select the checkpoint with the lowest validation error for color network fine-tuning. The color network is initialized with the weights of the fine NeRF network (used for depth dataset generation) and is trained on the RGB dataset for \(300,000\) iterations. The checkpoint with the lowest validation error is chosen for the final evaluation on the test set. During fine-tuning, the sampling network's parameters are frozen.

\begin{figure}[b]
  \centering

  % MIC
  \begin{subfigure}[b]{0.23\textwidth}
    \centering
    \includegraphics[trim=80px 200px 120px 60px, clip, width=\textwidth]{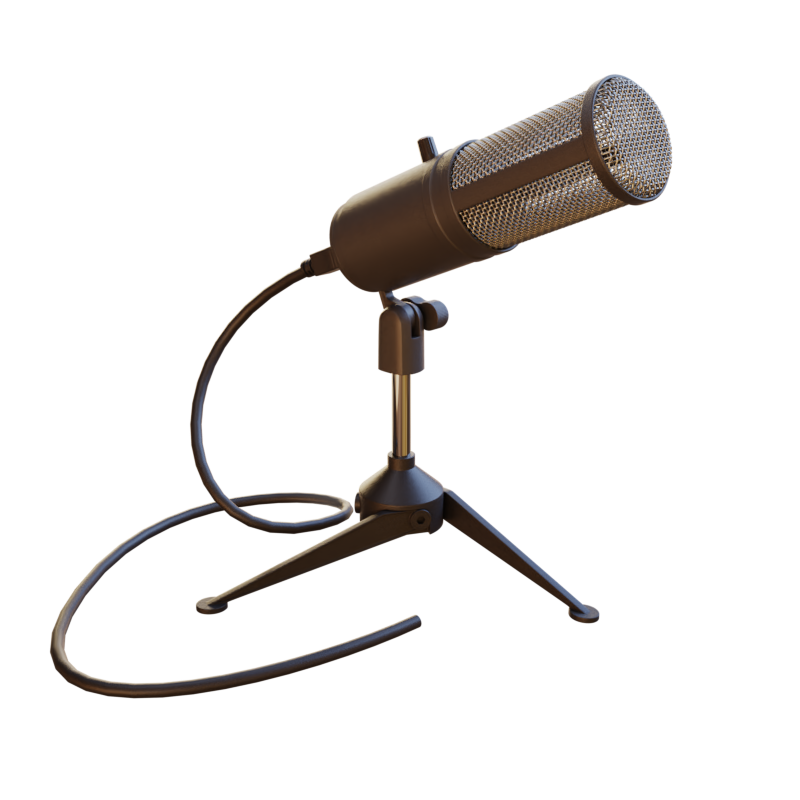}
    \caption{Ground Truth}
    \label{fig: experiments mic comparison gt}
  \end{subfigure}
  \begin{subfigure}[b]{0.23\textwidth}
    \centering
    \includegraphics[trim=40px 100px 60px 30px, clip, width=\textwidth]{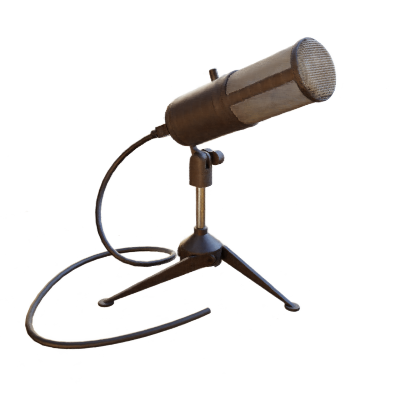}
    \caption{NeRF 64+128}
    \label{fig: experiments mic comparison nerf 128}
  \end{subfigure}
  \begin{subfigure}[b]{0.23\textwidth}
    \centering
    \includegraphics[trim=40px 100px 60px 30px, clip, width=\textwidth]{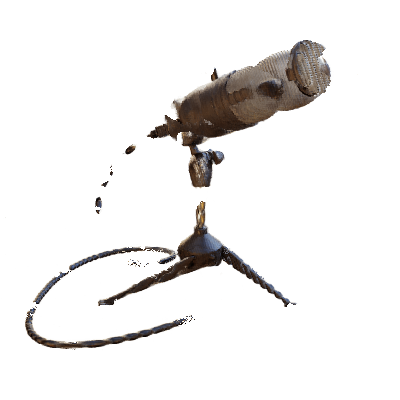}
    \caption{NeRF 8+16}
    \label{fig: experiments mic comparison nerf 16}
  \end{subfigure}
  \begin{subfigure}[b]{0.23\textwidth}
    \centering
    \includegraphics[trim=40px 100px 60px 30px, clip, width=\textwidth]{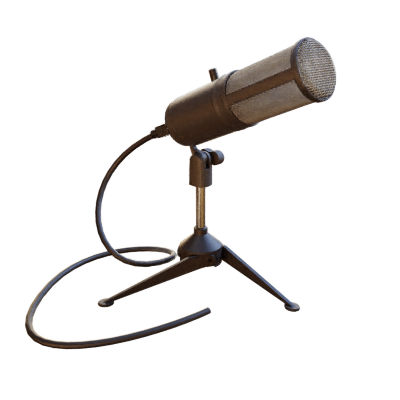}
    \caption{TermiNeRF 32}
    \label{fig: experiments mic comparison TermiNeRF}
  \end{subfigure}

  % SHIP
  \begin{subfigure}[b]{0.23\textwidth}
    \centering
    \includegraphics[trim=120px 240px 240px 200px, clip, width=\textwidth]{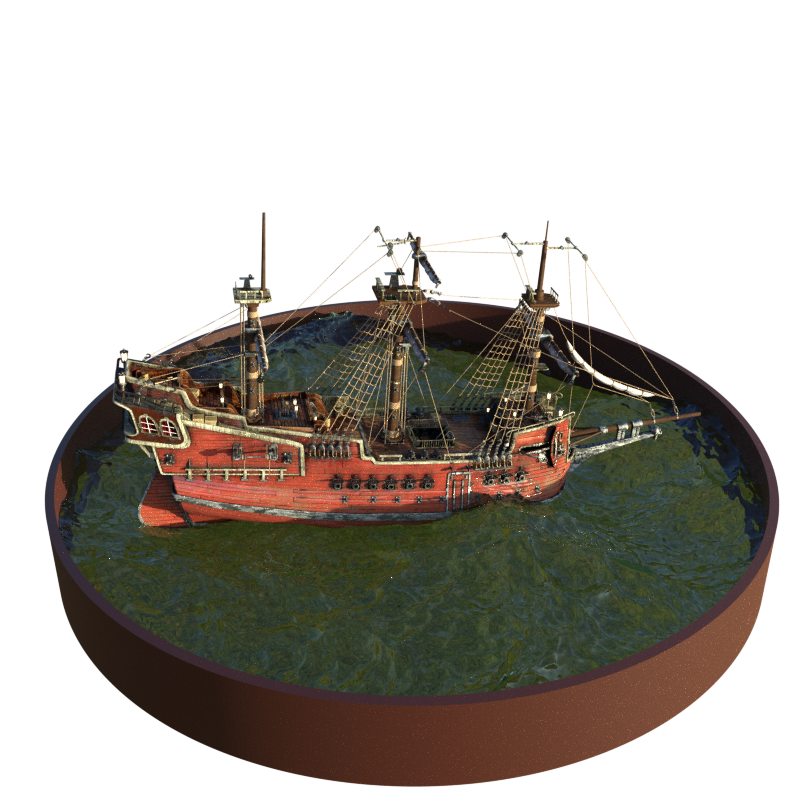}
    \caption{Ground Truth}
    \label{fig: experiments ship comparison gt}
  \end{subfigure}
  \begin{subfigure}[b]{0.23\textwidth}
    \centering
    \includegraphics[trim=60px 120px 120px 100px, clip, width=\textwidth]{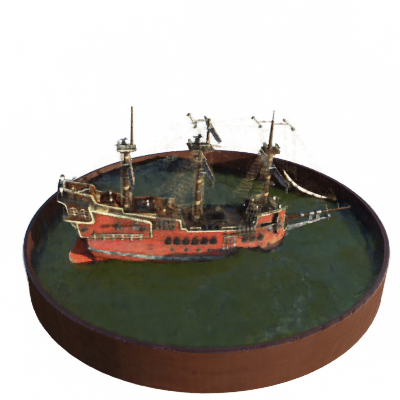}
    \caption{NeRF 64+128}
    \label{fig: experiments ship comparison nerf 128}
  \end{subfigure}
  \begin{subfigure}[b]{0.23\textwidth}
    \centering
    \includegraphics[trim=60px 120px 120px 100px, clip, width=\textwidth]{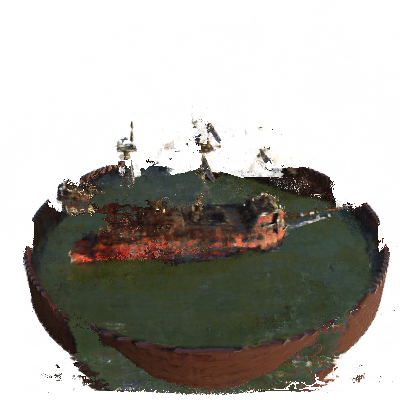}
    \caption{NeRF 8+16}
    \label{fig: experiments ship comparison nerf 16}
  \end{subfigure}
  \begin{subfigure}[b]{0.23\textwidth}
    \centering
    \includegraphics[trim=60px 120px 120px 100px, clip, width=\textwidth]{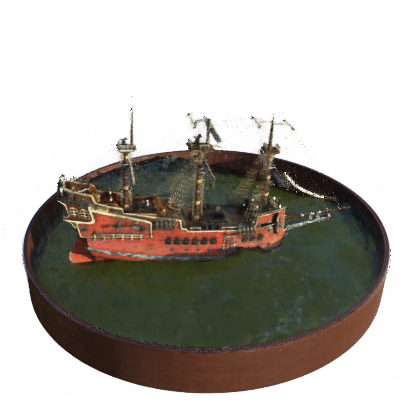}
    \caption{TermiNeRF 32}
    \label{fig: experiments ship comparison TermiNeRF}
  \end{subfigure}
  \hfill
  \caption{TermiNeRF and NeRF comparison}
  \label{fig: experiments TermiNeRF nerf comparison}
  \vspace{-6mm}
\end{figure}
For training of both networks, the Adam optimizer with the initial learning rate of \(5 \cdot 10^{-4}\) and \(5 \cdot 10^{-5}\) with decay was used for the sampling and the color networks.

\textbf{Results:}
\begin{table*}[h!]
  \centering
  \caption{Quantitative comparison of our method with NeRF. Although there is a marginal trade-off between the decreasing number of rendering samples and the image quality, both PSNR\(\uparrow\) and LPIPS\(\downarrow\) metrics values show how our method is able to render images faster, with the similar or higher-quality than NeRF. Each method also has a number that determines the number of rendering samples used.}
  \begin{tabular}{lcllllllllll} \toprule
    & Average
    & \multicolumn{2}{c}{Lego\cite{mildenhall2020nerf}}
    & \multicolumn{2}{c}{Drums\cite{mildenhall2020nerf}}
    & \multicolumn{2}{c}{Mic\cite{mildenhall2020nerf}}
    & \multicolumn{2}{c}{Ship\cite{mildenhall2020nerf}}
    & \multicolumn{2}{c}{Radiometer} \\

    Method & speed-up & PSNR & LPIPS & PSNR & LPIPS & PSNR & LPIPS & PSNR & LPIPS & PSNR & LPIPS \\\midrule
    NeRF 64+128 & 1\(\times\) & 37.04 & .0190 & 35.66 & .0542 & 39.89 & .0221 & 34.45 & .0965 & 36.84 & .0494 \\
    \midrule
    NeRF 16+32 & 4.01\(\times\) & 35.81 & .0715 & 35.32 & .1242 & 38.98 & .0633 & 33.80 & .1821 & 36.54 & .0776 \\
    TermiNeRF 64 & 3.79\(\times\) & \textbf{37.26} & \textbf{.0203} & \textbf{35.61} & \textbf{.0532} & \textbf{40.43} & \textbf{.0175} & \textbf{34.52} & \textbf{.0951} & \textbf{36.89} & \textbf{.0540} \\
    \midrule
    NeRF 8+16 & 7.99\(\times\) & 34.40 & .1792 & 34.92 & .2114 & 38.05 & .1450 & 33.00 & .2915 & 36.35 & .1342 \\
    TermiNeRF 32 & 7.32\(\times\) & \textbf{37.04} & \textbf{.0239} & \textbf{35.57} & \textbf{.0587} & \textbf{40.29} & \textbf{.0169} & \textbf{34.43} & \textbf{.1004} & \textbf{36.70} & \textbf{.0692} \\
    \midrule
    NeRF 4+8 & 21.1\(\times\) & 33.26 & .3797 & 34.36 & .4039 & 37.24 & .2965 & 31.68 & .4323 & 36.10 & .1416 \\
    TermiNeRF 16 & 13.49\(\times\) & \textbf{36.20} & \textbf{.0383} & \textbf{35.41} & \textbf{.0784} & \textbf{39.52} & \textbf{.0214} & \textbf{34.07} & \textbf{.1205} & \textbf{36.40} & \textbf{.0888} \\
    TermiNeRF 8 & 36.58\(\times\) & 34.23 & .1584 & 34.87 & .1714 & 37.99 & .0795 & 31.71 & .3477 & 36.14 & .1199 \\
  \end{tabular}
  \label{tab: experiments main psnr lpips}
  \vspace{-3mm}
\end{table*}
\Cref{fig: experiments TermiNeRF nerf comparison} shows how the TermiNeRF can maintain the image quality of the original NeRF with only a fraction of total network forward passes.
\Cref{fig: experiments mic comparison nerf 16,fig: experiments ship comparison nerf 16}
show how necessary a large number of samples is when we do not have any depth information. The number of samples 8+16 is chosen to match the total number (32) of forward passes through the networks (8 for the coarse network and 8+16 for the fine network) with the TermiNeRF 32.

\subsection{Joint training and fine-tuning}

\begin{figure}[h]
  \centering
  \includegraphics[width=0.95\linewidth]{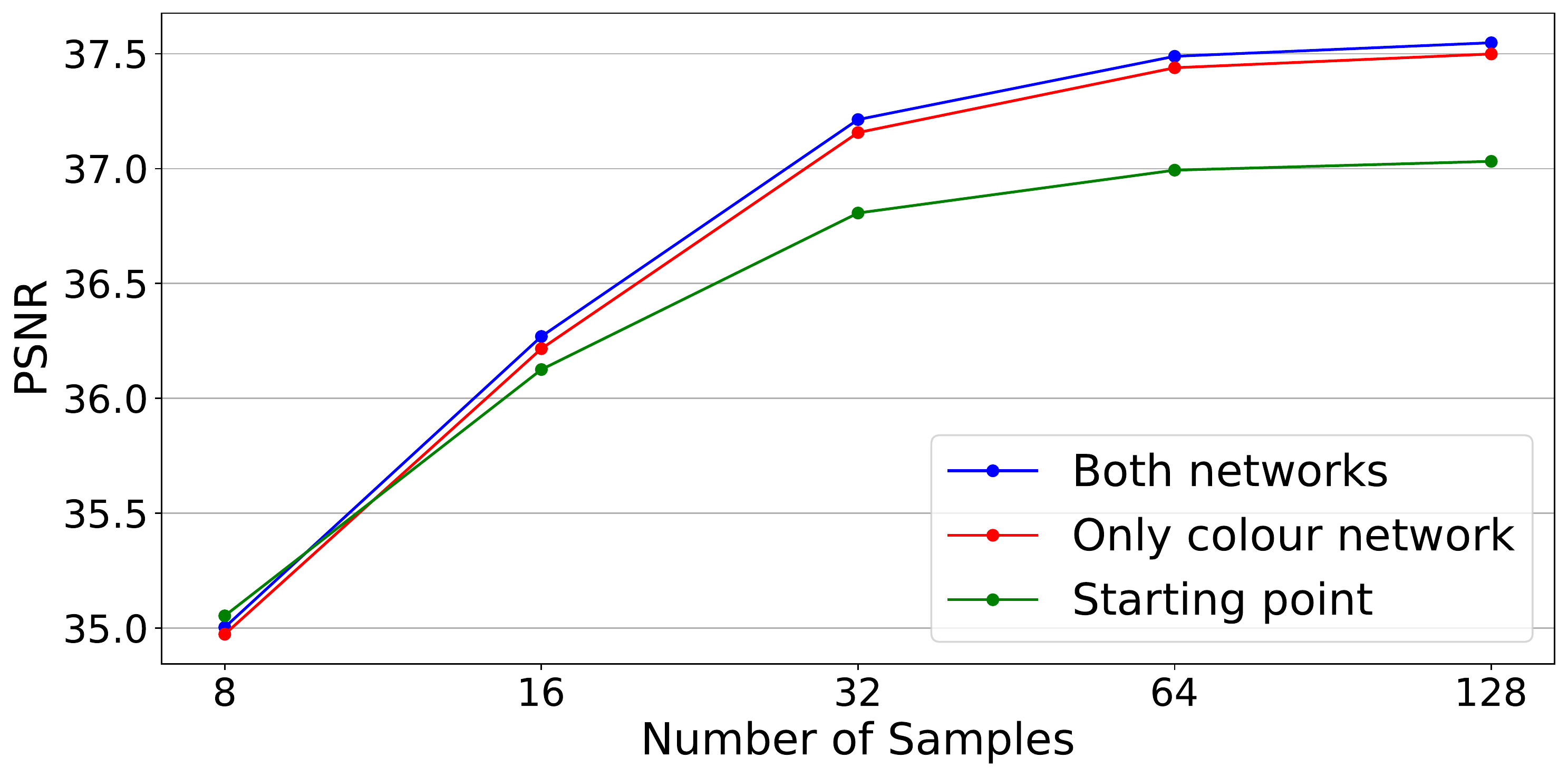}
  \caption{PSNR during fine tuning with and without optimizing the surface network}
  \label{fig: joint training psnr with samples}
\end{figure}

\begin{figure}[b]
  \def \factor {0.315}
  \def \vertspace {0cm}
  \def \vertspacebig {0.2cm}
  \def \horizontalspace {0.1cm}

  \centering
  \begin{tabular}{c@{\hspace{\horizontalspace}}c@{\hspace{\horizontalspace}}c}
    \includegraphics[trim=60px 220px 250px 90px, clip, width=\factor\linewidth]{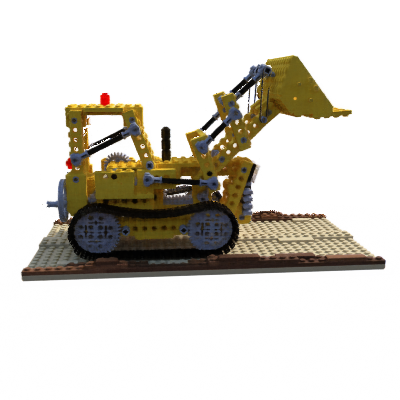} &
    \includegraphics[trim=60px 220px 250px 90px, clip, width=\factor\linewidth]{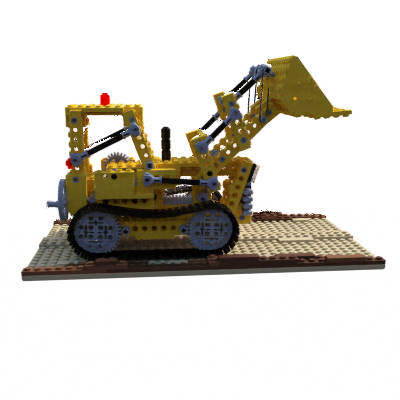} &
    \includegraphics[trim=60px 220px 250px 90px, clip, width=\factor\linewidth]{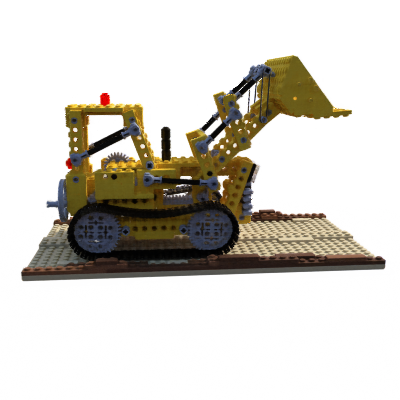} \vspace{\vertspace}\\

    \includegraphics[trim=115px 120px 135px 150px, clip, width=\factor\linewidth]{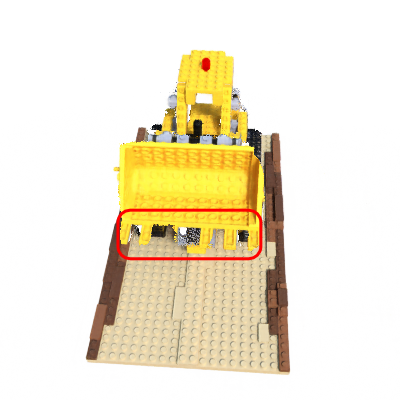} &
    \includegraphics[trim=115px 120px 135px 150px, clip, width=\factor\linewidth]{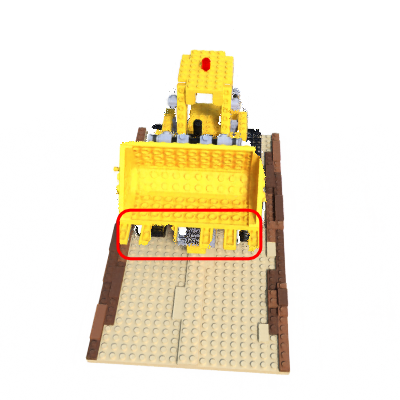} &
    \includegraphics[trim=115px 120px 135px 150px, clip, width=\factor\linewidth]{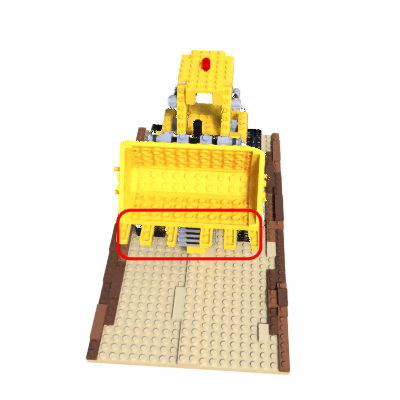} \vspace{\vertspace}\\

    \includegraphics[trim=0px 200px 200px 0px, clip, width=\factor\linewidth]{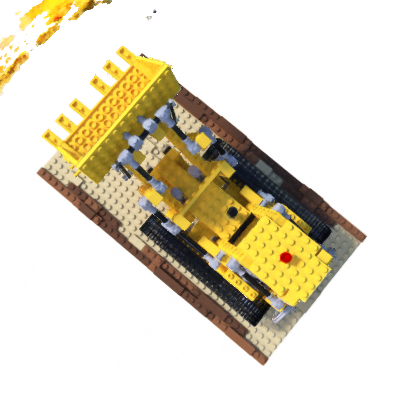} &
    \includegraphics[trim=0px 200px 200px 0px, clip, width=\factor\linewidth]{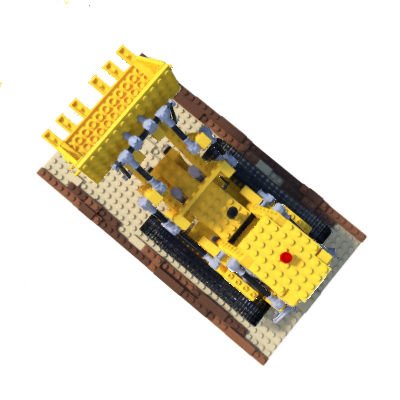} &
    \includegraphics[trim=0px 200px 200px 0px, clip, width=\factor\linewidth]{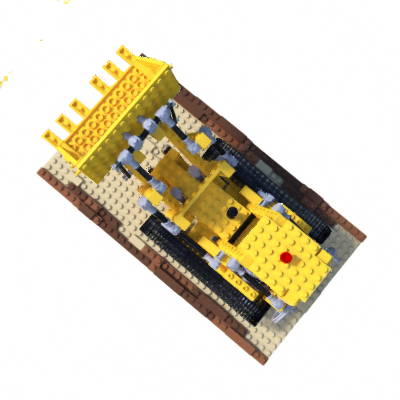} \vspace{\vertspace}\\

    Starting Point & Color Network & Both Networks\\
  \end{tabular}
  \caption{Joint Fine-Tuning: Example of improvements done by fine-tuning. We compare training both networks in alternating fashion with training only the color network.
  Images are chosen to demonstrate scenarios where a wrong color was rendered due to the surface misprediction, where fine-tuning the sampling network is necessary. For most general cases, such as the third row, further training with the sampling network frozen is sufficient.}
  \label{fig: experiments fine tuning comparison}
\end{figure}

The premise of this experiment is that as the depth dataset is obtained from the NeRF color network, the provided depth data should get better with fine-tuning of the network and, therefore, there is a potential for further training of the sampling network, too. This leads to a positive feedback loop where the sampling network gets more accurate with more precise depth data, and the color network learns to predict the color and volume distribution with better detail with more precisely placed samples.

The experiment set-up stays the same as the previous one, except that at the color network fine-tuning step, both color and sampling networks are updated in alternating iterations, where the depth supervision for the sampling network is provided by the color network. However, we can assign the opacity only to points evaluated with the color network. Assuming the points are suggested only by the predicted distribution, we would never acquire the true density value for regions of the ray that are left unsampled due to low weight, and the sampling network would never learn the additional geometry. To avoid this issue, for the joint fine-tuning, we sample 128 points from the predicted distribution with additional 64 equidistantly sampled points along the whole ray between \textit{near} and \textit{far} z-values.

\Cref{fig: joint training psnr with samples} shows that although fine-tuning itself is responsible for the majority of the improvement, we can still improve the image quality if we train the sampling network further.

A false-positive sampling network misprediction poses no issue for the color network, but the color network cannot fill in the gap caused by false-negative misprediction. Even though false-positives are occasional, they still happen and can be fixed only with more precise depth data and further training, which, in this case, happens jointly. \Cref{fig: experiments fine tuning comparison} focuses on image quality loss due to the sampling network issues.

It is important to note that even though we could train both networks jointly from the beginning, this process would be inefficient as the sampling network would be receiving mostly arbitrary training signal from the color network. Therefore, we first pre-train the color network. Once it converges, we train the sampling network and then jointly fine-tune both.

\begin{figure}[h]
  \def \factor {0.315}
  \def \vertspace {0cm}
  \def \vertspacebig {0.2cm}
  \def \horizontalspace {0.1cm}

  \centering
  \begin{tabular}{c@{\hspace{\horizontalspace}}c@{\hspace{\horizontalspace}}c}
    \includegraphics[trim=30px 70px 50px 50px, clip, width=\factor\linewidth]{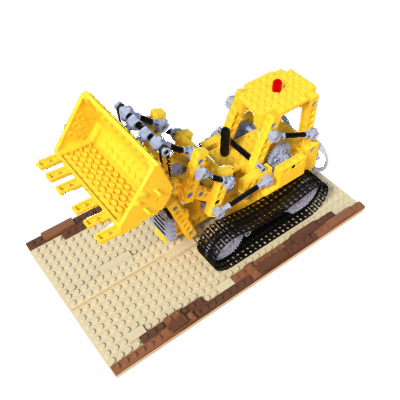} &
    \includegraphics[trim=30px 70px 50px 50px, clip, width=\factor\linewidth]{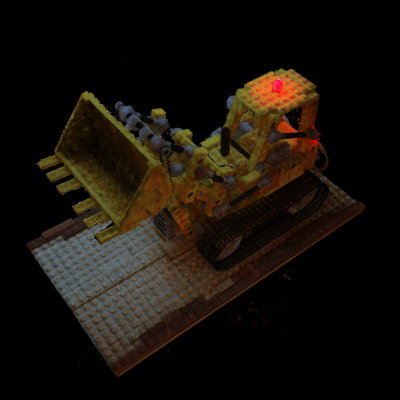} &
    \includegraphics[trim=30px 70px 50px 50px, clip, width=\factor\linewidth]{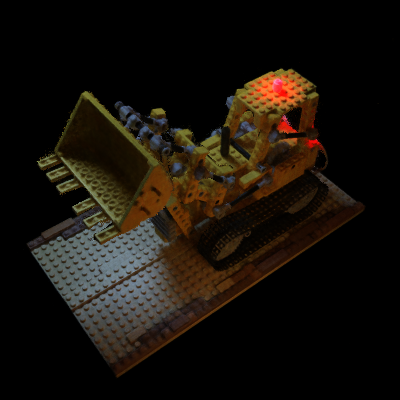}
     \vspace{\vertspace}\\

    \includegraphics[trim=30px 60px 30px 60px, clip, width=\factor\linewidth]{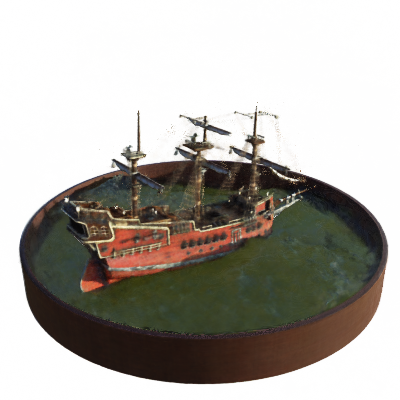} &
    \includegraphics[trim=30px 60px 30px 60px, clip, width=\factor\linewidth]{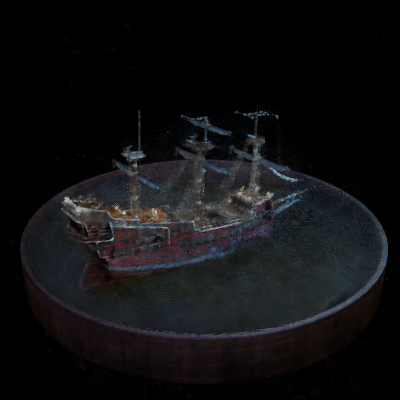} &
    \includegraphics[trim=30px 60px 30px 60px, clip, width=\factor\linewidth]{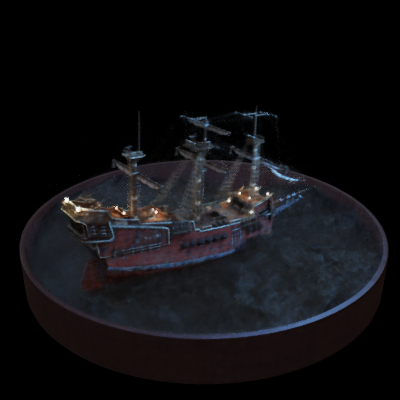}
    \vspace{\vertspace}\\

    Initial & Adapted (3m) & Adapted (1h)\\
  \end{tabular}
  \caption{Renders of the adapted network to edited scenes. The respective training times on GTX1080 are included for both renders of the adapted scene.}
  \label{fig: experiments modified scene comparison}
\end{figure}

\textbf{Fine-tuning for modified scene:}
Without prior depth information, during the training each ray has to be sampled along its full length from \textit{near} to \textit{far} boundaries. This process is just as inefficient as rendering. However, once the sampling network has been trained on one object, the color network can be quickly (\cref{fig: experiments modified scene comparison}) retrained for a new scene that shares the same geometric structure, e.g., for a scene with different lighting conditions or different colors or textures.

\begin{figure}[h]
  \centering
  \includegraphics[width=0.95\linewidth]{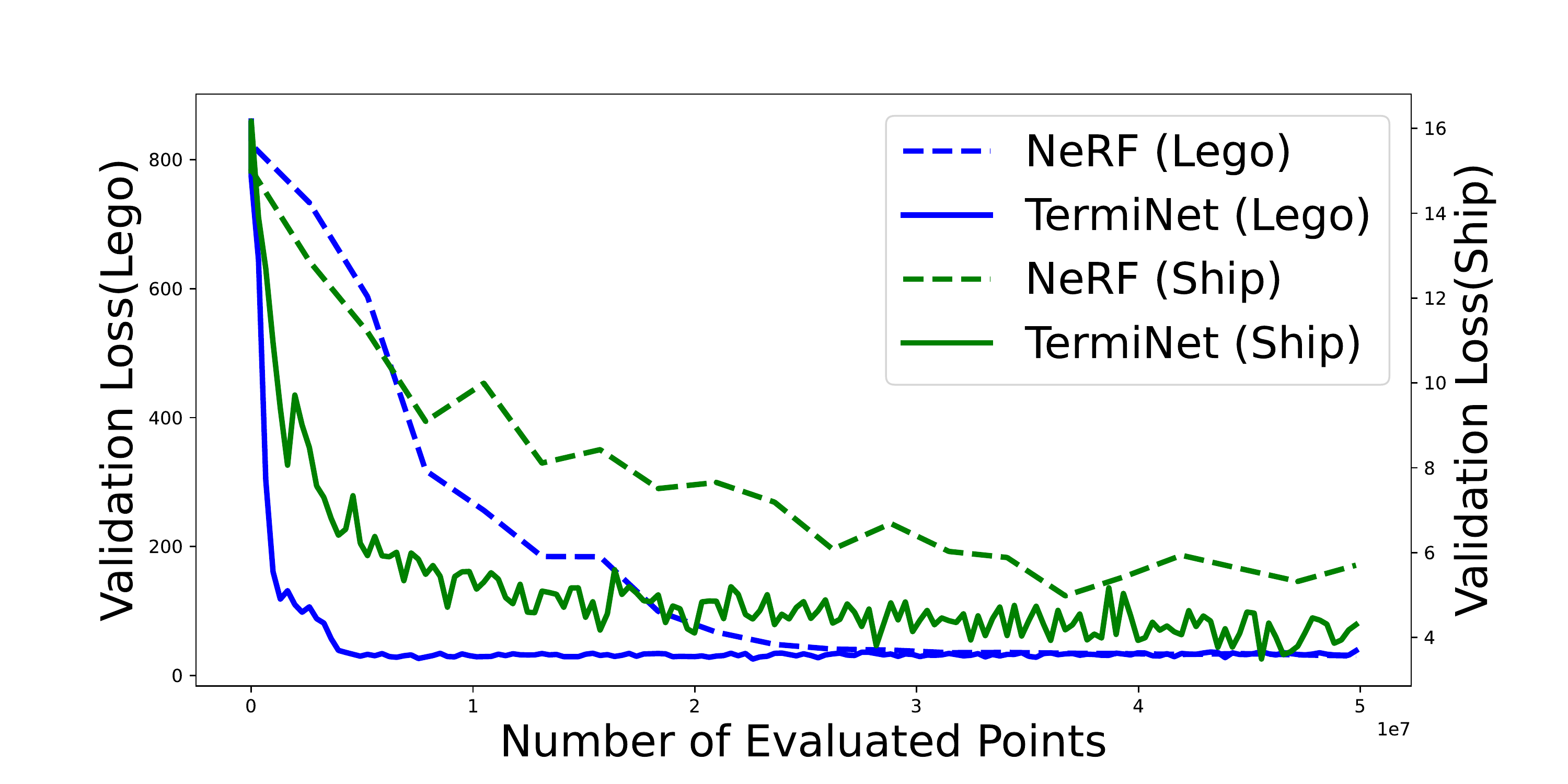}
  \caption{The training time is proportional to the number of necessary forward passes until the training converges. The color network trained with better places samples is able to converge to the solution with fewer evaluated points. This, in turn, results in shorter training times and faster adaptation to a modified dataset.}
  \label{fig: experiments psnr modified scene}
\end{figure}

% \FloatBarrier

\subsection{Comparison with other methods}
We provide a comparison of our work to the original NeRF implementation, DONeRF \cite{neff2021donerf} and other related methods \cite{autoint,kilonerf,Rebain_2021_CVPR,yu2021plenoctrees}. A more in-depth comparison with NeRF can be found in the \cref{tab: experiments main psnr lpips}.

\textbf{DONeRF comparison details:}
For comparison, we use our own implementation of DONeRF \cite{neff2021donerf} (as no official code is available), which we refer to as DONeRF*.

%Naturally, by capturing most objects, we obtain high-frequency depth maps with values changing rapidly near the edges of the object. Estimating these values with a continuous function approximated by a neural network poses an issue.

We replicate DONeRF's \cite{neff2021donerf} discretization where we first split up the ray into \(N\) bins and then discretize the depth value \(d_s\) along the ray and assign the classification value \(C_{x, y}(z)\) to bin \(z\) of the ray \((x, y)\) as
\begin{equation}
  C = \begin{cases}
    1  & \quad \text{if } d_z \leq d_s < d_{z+1} \\
    0  & \quad \text{otherwise}
  \end{cases}
\end{equation}

In addition, we replicate the DONeRF's blurring approach, where each ray is blurred as follows:
\begin{equation}
  \hat{C}_{x, y}(z) = \max_{i, j \in \pm \lfloor \rfrac{K}{2} \rfloor} \left( C_{x+i, y+j}(z) - \frac{\sqrt{i^2 + j^2}}{\sqrt(2) \lfloor \rfrac{K}{2} \rfloor} \right)
\end{equation}
and filtering
\begin{equation}
  \grave{C}_{x, y}(z) = \min\left(\sum_{i=-\lfloor \rfrac{Z}{2} \rfloor}^{\lfloor \rfrac{Z}{2} \rfloor} \hat{C}_{x, y}(z+i) \frac{\lfloor \rfrac{Z}{2} \rfloor + 1 - |i|}{\lfloor \rfrac{Z}{2} \rfloor + 1}  \right)
\end{equation}
Therefore our re-implementation follows exactly the original DONeRF method, the only difference being that as we are training on \(360^\circ\) datasets, and the DONeRF focused on forward facing scenes, we omit the logarithmic sampling and warping from the implementation as we found these do not perform well on \(360^\circ\) scenes. The input to the network are therefore equidistantly sampled points and the ray segments (bins) are of a constant length.

\begin{figure}[h]
  \def \factor {0.315}
  \def \vertspace {0cm}
  \def \vertspacebig {0.2cm}
  \def \horizontalspace {0.1cm}

  \centering
  \begin{tabular}{c@{\hspace{\horizontalspace}}c@{\hspace{\horizontalspace}}c}
    \includegraphics[trim=130px 160px 130px 100px, clip, width=\factor\linewidth]{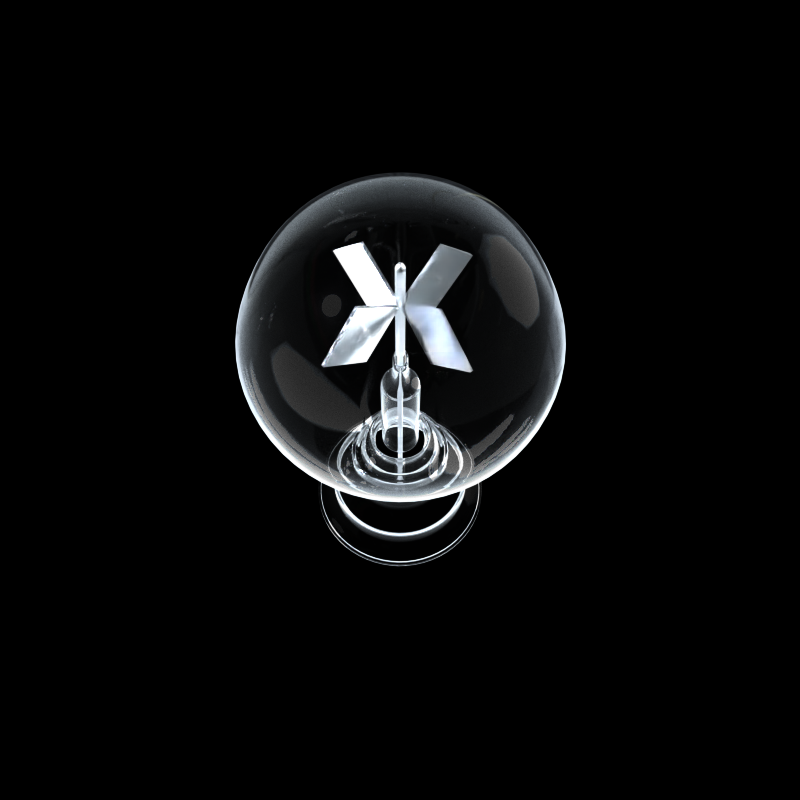} &
    \includegraphics[trim=65px 80px 65px 50px, clip, width=\factor\linewidth]{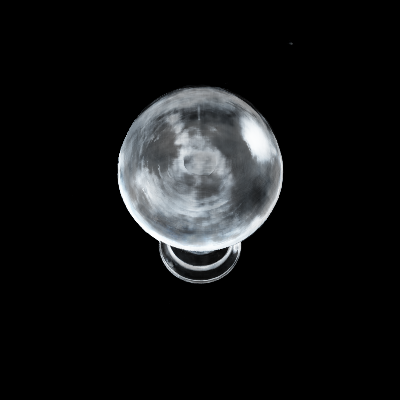} &
    \includegraphics[trim=65px 80px 65px 50px, clip, width=\factor\linewidth]{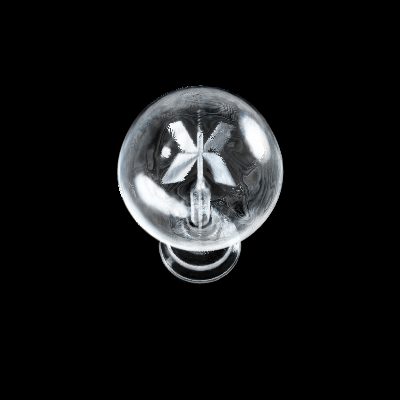} \vspace{\vertspace}\\

    \includegraphics[trim=130px 130px 130px 130px, clip, width=\factor\linewidth]{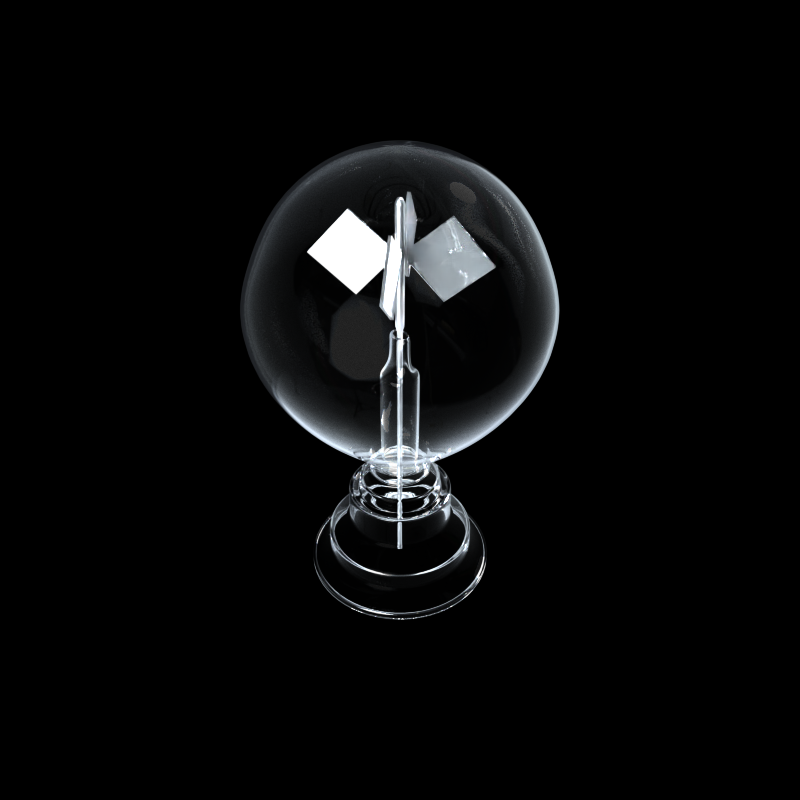} &
    \includegraphics[trim=65px 65px 65px 65px, clip, width=\factor\linewidth]{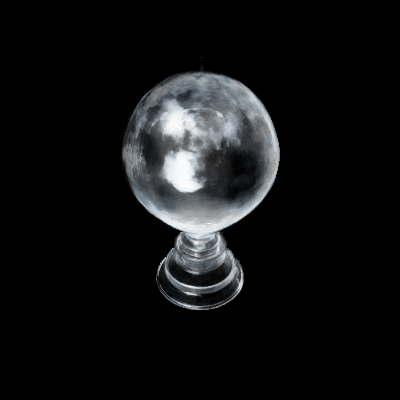} &
    \includegraphics[trim=65px 65px 65px 65px, clip, width=\factor\linewidth]{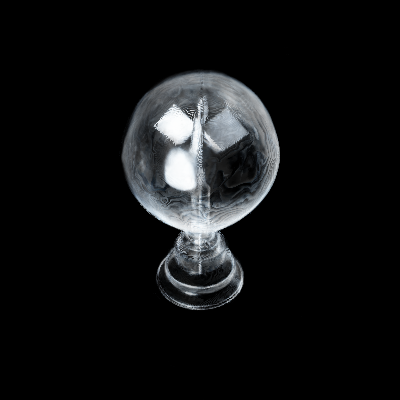} \vspace{\vertspace}\\

    Ground Truth & DONeRF* 32 & TermiNeRF 32\\
  \end{tabular}
  \caption{Qualitative comparison of renders produced by depth network trained on ground truth (GT) depths in the DONeRF style and sampling network trained on NeRF weights. A single depth value is not enough to capture multiple layers of volume. All images are from the test set and were rendered with 32 samples drawn from the distributions predicted by the sampling network.}
  \label{fig: ground truth depth comparison}
\end{figure}

\begin{figure}[h!]
  \centering
    \includegraphics[width=1.0\linewidth]{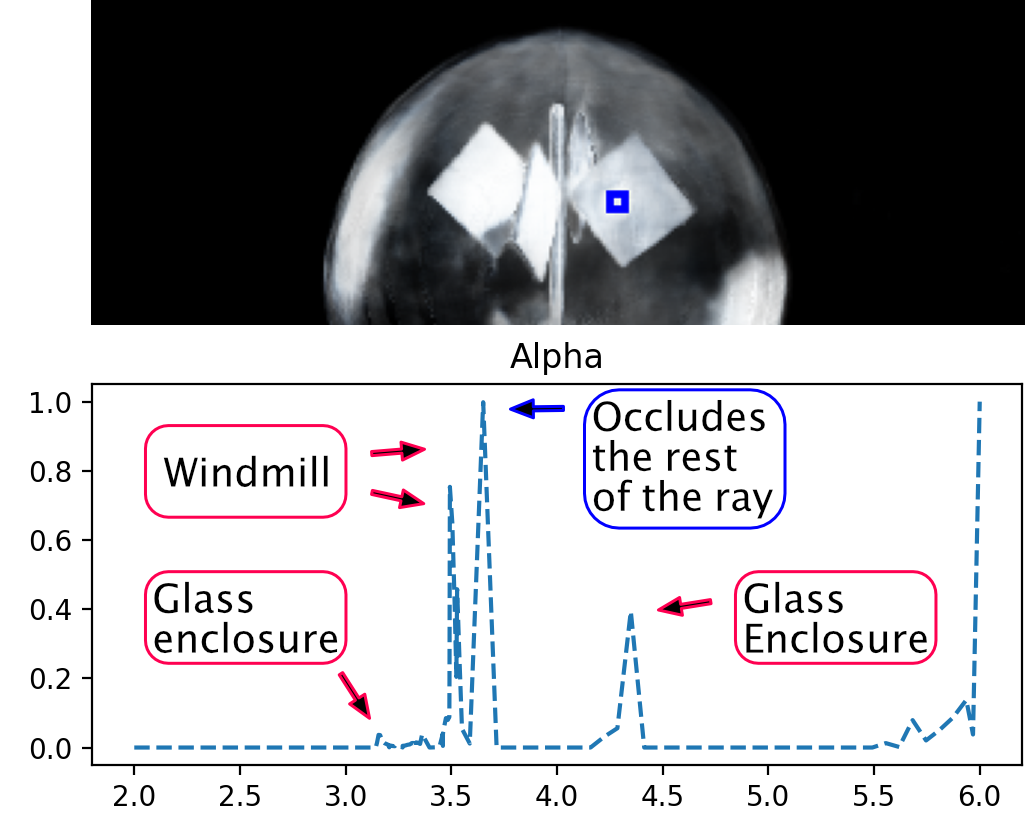}
    \caption{Density of the points at various z-values along the ray cast through the center of the blue square. Multiple peaks show different surfaces hit by the ray: glass, windmill, glass and background.}
  \end{figure}

Although ground truth depth maps are very precise, the single depth value per ray ultimately lacks the expressivity to capture a more complicated volume that contains multiple layers of depth, such as glass or fog. Such shortcoming can be seen on \cref{fig: ground truth depth comparison} where the whole distribution of ray-volume intersections are necessary to capture the inside of the radiometer.

Additionally, our method does not model reflections, refractions, or any other effects where the direction of a ray deviates from a straight line. Even though some of the effects can be modeled through the view direction-dependent component of the input, reflections like in the \cref{fig: ground truth depth comparison} can sometimes be over-expressed in the final render.

\textbf{Numerical comparison:}
In this section, we put TermiNeRF into perspective and provide numerical results and comparisons with our best effort DONeRF* implementation, original NeRF\cite{mildenhall2020nerf}, KiloNeRF\cite{kilonerf}, PlenOctrees\cite{yu2021plenoctrees} and Autoint\cite{autoint} on \(360^\circ\) Lego and Ship datasets.

\begin{figure}[h]
  \centering
  \includegraphics[width=0.94\linewidth]{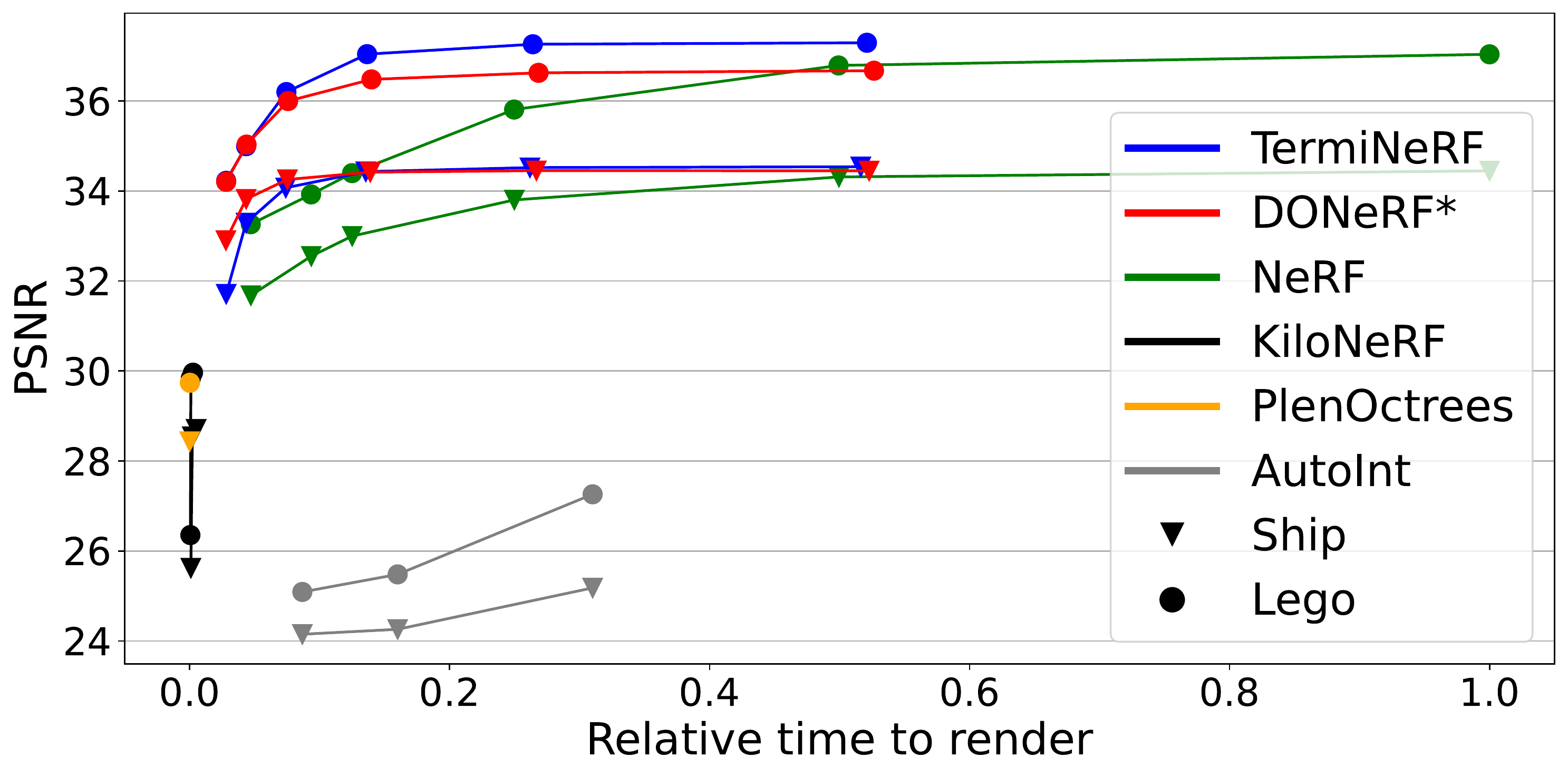}
  \caption{PSNR comparison plot for Ship and Lego}
  \label{fig: comparison psnr plot}
\end{figure}

\begin{figure}[h]
  \centering
  \includegraphics[width=0.94\linewidth]{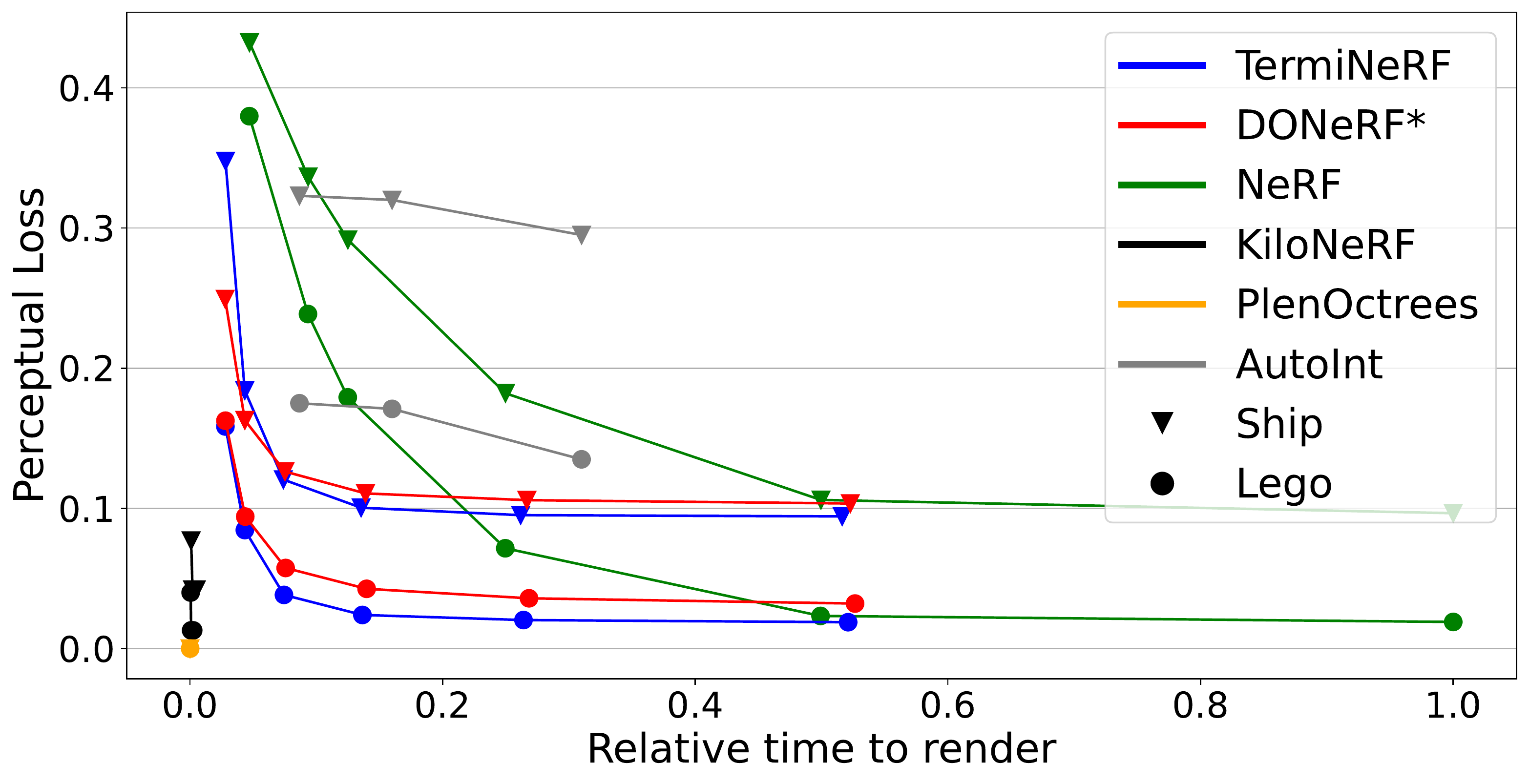}
  \caption{LPIPS comparison plot for Ship and Lego}
  \label{fig: comparison lpips plot}
\end{figure}

\section{Conclusions}
Neural Radiance Fields produce high-quality renders but are computationally expensive to render. In this paper, we propose a \textit{sampling network} to focus only on the regions of a ray that yield a color, which effectively allows us to decrease the number of necessary forward passes through the network, speeding up the rendering pipeline \(\approx 14 \times\).
The network performs one-shot ray-volume intersection distribution estimations, keeping the pipeline efficient and fast. We use only RGB data to train the model, which makes this method not only versatile, but also more accurate and suitable for translucent surfaces.

Even though there are methods that render significantly faster than our approach, the benefits of our method are not limited to rendering/inference but also are applicable during training. For example, TermiNeRF can be trained end-to-end and offers shorter training times when fine-tuning scenes. As we show, this makes our method ideal for quickly adapting to an edited scene (for example, changing the texture, color, lighting or minor geometry modifications).

%-------------------------------------------------------------------------

\FloatBarrier
% \subsection{References}

{\small
\bibliographystyle{ieee_fullname}
\bibliography{egbib}
}

\end{document}